\documentclass[journal]{IEEEtran}
\usepackage{amsmath,amsfonts}
\usepackage{bbm}
\usepackage{url}
\usepackage{arydshln}
\usepackage{booktabs}
\usepackage[flushleft]{threeparttable}
\usepackage{color}
\usepackage{listings}

\ifCLASSINFOpdf
  \usepackage[pdftex]{graphicx}
\else
  \usepackage[dvips]{graphicx}
\fi
\begin{document}
%

\title{Physics-Informed Deep Learning for \\ Traffic State Estimation}

%
%
%

\author{Rongye~Shi,~\IEEEmembership{Member,~IEEE,}
        Zhaobin Mo, Kuang Huang, Xuan Di,~\IEEEmembership{Member,~IEEE,}
        and~Qiang~Du
\thanks{Manuscript received month day, year; revised month day, year.}
\thanks{This work was supported by the Data Science Institute at Columbia University under the Seed Funds Program. \textit{(Corresponding author: Rongye Shi and Xuan Di.)}}
\thanks{Rongye Shi and Zhaobin Mo are with the Department of Civil Engineering and Engineering Mechanics, Columbia University, New York, NY, 10027 USA (e-mail: rongye.shi@columbia.edu, zm2302@columbia.edu).}
\thanks{Kuang Huang is with the Department of Applied Physics and Applied Mathematics, Columbia University, New York, NY, 10027 USA (e-mail: kh2862@columbia.edu).}
\thanks{Xuan Di is with the Department of Civil Engineering and Engineering Mechanics, Columbia University, New York, NY, 10027 USA, and also with the Data Science Institute, Columbia University, New York, NY, 10027 USA (e-mail: sharon.di@columbia.edu).}
\thanks{Qiang Du is with the Department of Applied Physics and Applied Mathematics, Columbia University, New York, NY, 10027 USA, and also with the Data Science Institute, Columbia University, New York, NY, 10027 USA (e-mail: qd2125@columbia.edu).}
}

%
%

\markboth{}%
{Shi \MakeLowercase{\textit{et al.}}: Physics-Informed Deep Learning for Traffic State Estimation}
%



\maketitle

\begin{abstract}

Traffic state estimation (TSE), which reconstructs the traffic variables (e.g., density) on road segments using partially observed data, plays an important role on efficient traffic control and operation that intelligent transportation systems (ITS) need to provide to people.  Over decades, TSE approaches bifurcate into two main categories, model-driven approaches and data-driven approaches. However, each of them has limitations: the former highly relies on existing physical traffic flow models, such as Lighthill-Whitham-Richards (LWR) models, which may only capture limited dynamics of real-world traffic, resulting in low-quality estimation, while the latter requires massive data in order to perform accurate and generalizable estimation. To mitigate the limitations, this paper introduces a physics-informed deep learning (PIDL) framework to efficiently conduct high-quality TSE with small amounts of observed data. PIDL contains both model-driven and data-driven components, making possible the integration of the strong points of both approaches while overcoming the shortcomings of either. This paper focuses on highway TSE with observed data from loop detectors, using traffic density as the traffic variables. We demonstrate the use of PIDL to solve (with data from loop detectors) two popular physical traffic flow models, i.e., Greenshields-based LWR and three-parameter-based LWR, and discover the model parameters. We then evaluate the PIDL-based highway TSE using the Next Generation SIMulation (NGSIM) dataset. The experimental results show the advantages of the PIDL-based approach in terms of estimation accuracy and data efficiency over advanced baseline TSE methods.

\end{abstract}


%
\IEEEpeerreviewmaketitle

\section{Introduction}
\IEEEPARstart{T}{traffic} state estimation (TSE) is one of the central components that supports traffic control, operations and other transportation services that intelligent transportation systems (ITS) need to provide to people with mobility needs. For example, the operations and planning of intelligent ramp metering, and traffic congestion management rely on overall understanding of road-network states. However, only a subset of traffic states can be observed using traffic sensing systems deployed on roads or probe vehicles moving along with the traffic flow. To obtain the overall traffic state profile from limited observed information defines the TSE problem. Formally, TSE refers to the data mining problem of reconstructing traffic state variables, including but not limited to flow (veh/h), density (veh/km), and speed (km/h), on road segments using partially observed data from traffic sensors~\cite{Seo-17}. 

TSE research can be dated back to late 1940s~\cite{Berry-1949} in existing literature and has continuously gained great attention in recent decades.  TSE approaches can be briefly divided into two main categories: model-driven and data-driven. A model-driven approach is based on a priori knowledge of traffic dynamics, usually described by a physical model, e.g., the Lighthill-Whitham-Richards (LWR) model~\cite{Lighthill-1955, Richards-1956}, to estimate the traffic state using partial observation. It assumes the model to be representative of the real-world traffic dynamics such that the unobserved values can be properly added using the model with less input data. The disadvantage is that existing models, which are provided by different modelers, may only capture limited dynamics of real-world traffic,  resulting in low-quality estimation in the case of inappropriately-chosen models and poor model calibrations. Paradoxically, sometimes, verifying or calibrating a model requires a large amount of observed data, undermining the data efficiency of model-driven approaches. 

A data-driven approach is to infer traffic states based on the dependence learned from historical data using statistical or machine learning (ML) methods. Approaches of this type do not use any explicit traffic models or other theoretical assumptions, and can be treated as a ``black box" with no interpretable and deductive insights. The disadvantage is that in order to maintain a good generalizable inference to long-term unobserved values, massive and representative historical data are a prerequisite, leading to high demands on data collection infrastructure and enormous installation-maintenance costs.  

To mitigate the limitations of the previous TSE approaches, this paper, for the first time (to the best of our knowledge), introduces a framework, physics-informed deep learning (PIDL), to the TSE problem for  improved estimation accuracy and data efficiency. The PIDL framework is originally proposed for solving nonlinear partial differential equations (PDEs) in recent years~\cite{Raissi-2018a,  Raissi-2018b}. PIDL contains both a model-driven component (a physics-informed neural network for regularization) and a data-driven component (a physics-uninformed neural network for estimation), making possible the integration of the  strong  points  of both model-driven and data-driven approaches while overcoming the weaknesses of either. This paper focuses on demonstrating PIDL for TSE on highways using observed data from inductive loop detectors, which are the most widely used sensors. We will show the benefit of using the LWR-based physics to inform  deep learning. Grounded on the highway and loop detector setting, we made the following contributions: We

\vspace{2mm}
\begin{itemize}

\item Establish a PIDL framework for TSE, containing a physics-uninformed neural network  for estimation and a physics-informed neural network  for regularization. Specifically, the  latter  can  encode  a  traffic  flow  model, which  will  have  a  regularization  effect  on  the  former, making its estimation physically consistent;

\vspace{2mm}

\item Design two PIDL architectures for TSE with traffic dynamics described by Greenshields-based LWR and three-parameter-based LWR, respectively. We show the ability of PIDL-based method to estimate the traffic density in the spatio-temporal field of interest using initial data or limited observed data from loop detectors, and in addition, to discover model parameters;

\vspace{2mm}

\item Demonstrate the robustness of PIDL using real data, i.e., the data of a highway scenario in the Next Generation  SIMulation (NGSIM) dataset. The experimental results show the advantage of PIDL in terms of estimation accuracy and data efficiency over TSE baselines, i.e., an extended Kalman filter (EKF), a neural network, and a long short-term memory (LSTM) based method.

\vspace{2mm}

\end{itemize}

The rest of this paper is organized as follows.  Section II briefs related work on TSE and PIDL. Section III formalizes the PIDL framework for TSE. Sections IV and V detail the designs and experiments of PIDL for Greenshields-based LWR and three-parameter-based LWR, respectively. Section~VI evaluates the PIDL on NGSIM data over baselines. Section VII concludes our work.

\section{Related Work}

\subsection{Traffic State Estimation Approaches}

A number of studies tackling the problem of TSE have been published in recent decades. As discussed in the previous section, this paper briefly divides TSE approaches into two main categories: model-driven and data-driven. For more comprehensive categorization, we refer readers to~\cite{Seo-17}.

Model-driven approaches accomplish their estimation process relying on traffic flow models, such as the LWR~\cite{Lighthill-1955, Richards-1956}, Payne-Whitham (PW)~\cite{Payne-1971, Whitham-1974}, Aw-Rascle-Zhang (ARZ)~\cite{Aw-2002, zhang-2002}  models for one-dimensional traffic flow (link models) and junction models~\cite{Garavello-2016} for network traffic flow (node models). Most estimation approaches in this category are data assimilation (DA) based, which attempt to  find ``the most likely state", allowing observation to correct the model's prediction. Popular examples include the Kalman filter (KF) and its variants (e.g., extended KF~\cite{Yibing-2008, Yibing-2009, xuan-2010}, unscented KF~\cite{Mihaylova-2006}, ensemble KF~\cite{Blandin-2012}), which find the state that maximizes the conditional probability of the next state given current estimate. Other than KF-like methods, particle filter (PF)~\cite{Mihaylova-2004} with improved  nonlinear representation, adaptive smoothing filter (ASF)~\cite{Treiber-2002} for combining multiple sensing data, were proposed to improve and extend different aspects of the TSE process.
In addition to DA-based methods, there have been many studies utilizing microscopic trajectory models to simulate the state or vehicle trajectories given some boundary conditions from data~\cite{Coifman-2002,laval-2012,Kuwahara-2013, Blandin-2013, Fan-2014, Kawai-2019, Jabari-2019}. 
Among model-driven approaches, LWR model is of the most popular deterministic traffic flow models due to its relative simplicity, compactness, and success in capturing real-world traffic phenomena such as shockwave~\cite{Seo-17}. 
\textcolor{black}{To perform filtering based statistical methods on an LWR model, we usually need to make a prior assumption about the distribution of traffic states. The existing practice includes two ways: one to add a Brownian motion on the top of deterministic traffic flow models, leading to Gaussian stochastic traffic flow models. 
and the other to derive intrinsic stochastic traffic flow models with more complex probabilistic distributions \cite{paveri1975boltzmann,davis1994estimating,kang1995estimation,xuan-2010,jabari2012stochastic}. 
In this paper, we will use the stochastic LWR model with Gaussian noise (solved by EKF) as one baseline to validate the superior performance of our model.
}

Data-driven approaches estimate traffic states using historical data or streaming data without explicit traffic flow models. Early studies considered relatively simple statistical methods, such as heuristic imputation techniques using empirical statistics of historical data~\cite{Smith-2003}, linear regression model~\cite{Chen-2003}, and autoregressive moving average (ARMA) model for time series data analysis~\cite{Zhong-2004}. To handle more complicated traffic data, ML approaches were involved, including principal component analysis (PCA) and its variants~\cite{Li-2014, Tan-2014}, k-nearest neighbors (kNN)~\cite{Tak-2016}, and probabilistic graphical models (i.e., Bayesian networks)~\cite{Ni-2005}. Deep learning model~\cite{Polson-2017}, long short term memory (LSTM) model~\cite{LiWei-2018}, deep embedded model~\cite{ZhengZibin-2019} and fuzzy neural network (FNN)~\cite{Tang-2020} have recently been applied for traffic flow prediction and imputation.

Each of these two approaches has disadvantages, and it is promising to develop a framework to integrate physics to ML, such that we can combine the  strong  points of both model-driven and data-driven approaches while overcoming the weaknesses of either. This direction has gained increasing interest in recent years. Hofleitner \textit{et al}~\cite{Hofleitner-2012} and Wang \textit{et al}~\cite{Wang-2019} developed hybrid models for estimating travel time on links for the signalized intersection scenarios, combining Bayesian network and hydrodynamic theory. 
\textcolor{black}{
Jia \textit{et al}~\cite{Xiaowei-2019} developed physics guided recurrent neural networks that learn from calibrated model predictions in addition to real data to model water temperature in lakes. Yuan \textit{et al}~\cite{YuanY-2020} recently proposed to leverage a hybrid framework, physics regularized Gaussian process (PRGP)~\cite{WangZ-2020} for macroscopic traffic flow modeling and TSE.} Our paper contributes to the trend of developing hybrid methods for transportation research, and explores the use of the PIDL framework to addressing the TSE problem.

\subsection{Physics-Informed Deep Learning for Solving PDEs}

The concept of PIDL is firstly proposed by Raissi in 2018 as an effective alternative PDE solver to numerical schemes~\cite{Raissi-2018a, Raissi-2018b}. It approximates the unknown nonlinear dynamics governed by a PDE with two deep neural networks: one for predicting the unknown solution and the other, in which the PDE is encoded, for evaluating whether the prediction is consistent with the given physics. Increasing attentions have been paid to the application of PIDL in scientific and engineering areas, to name a few, the discovery of constitutive laws for flow through porous media~\cite{Yang-2019}, the prediction of vortex-induced vibrations~\cite{Maziar-2019}, 3D surface reconstruction~\cite{Fang-2020}, and the inference of hemodynamics in intracranial aneurysm~\cite{Maziar-2020}. In addition to PDEs, researchers have extended PIDL to solving space fractional differential equations~\cite{Gulian-2019} and space-time fractional advection-diffusion equations~\cite{Pang-2019}.

\subsection{Motivations of PIDL for TSE}

\begin{figure}[h!]
\centering
  \includegraphics[scale=0.8]{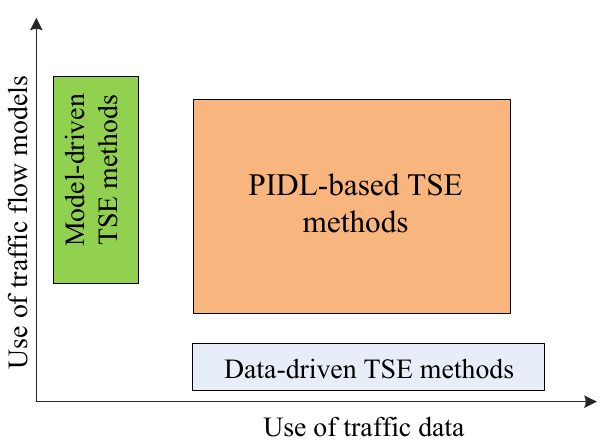}
  \caption{A presentation  of data-driven, model-drive, and PIDL-based TSE approaches in a 2D view, where $x$-axis is the use of traffic data and $y$-axis is the use of existing traffic flow models. (adapted from~\cite{Karpatne-2017})
}
  \label{fig:ch2-diagram}
\end{figure}

This paper introduces the framework of PIDL to  TSE problems. To be specific, we develop PIDL-based approaches \textit{to estimate the spatio-temporal traffic density of a highway segment over a period of time using observation from a limited number of loop detectors}. As shown in Fig.~\ref{fig:ch2-diagram}, in contrast to model-driven and data-driven approaches, the proposed PIDL-based TSE approaches are able to combine the advantages of both approaches by makeing efficient use of traffic data and existing knowledge in traffic flow models. As real-world traffic is composed of both physical and random human behavioral components, a combination of model-driven and data-driven approaches has a great potential to handle such complex dynamics.

A PIDL-based TSE approach consists of one deep neural network to estimate traffic state (data-driven component), while encoding the traffic models into another neural network to regularize the estimation for physical consistency (model-driven component). A PIDL-based approach also has the ability to discover unknown model parameters, which is an important problem in transportation research. To the best of our knowledge, this is the first work of combining  traffic flow models and deep learning paradigms for TSE.

\section{PIDL Framework Establishment for TSE}

This section introduces the PIDL framework in the context of TSE at a high-level, and Section~\ref{sec-IV}, \ref{sec-V}, \ref{sec-VI} will flesh out the framework for specific traffic flow models and their corresponding TSE problems.

The PIDL framework, which consists of a physics-uninformed neural network following by a physics-informed neural network, can be used for (1) approximating traffic states described by some traffic flow models in the form of nonlinear differential equations and (2) discovering unknown model parameters using observed traffic data.

\subsection{PIDL for TSE}
\label{3-A}

Let $\mathcal{N}[\cdot]$ be a general nonlinear differential operator and $\Omega$ be a subset of $\mathbb{R}^d$. For one lane TSE, $d$ is one by default in this paper, i.e., $\Omega = [0,L], L\in \mathbb{R}^+$. Then, the problem is to find the traffic state $\rho(t,x)$ at each point $(t,x)$ in a continuous domain, such that the following PDE of a traffic flow model can be satisfied:

\begin{equation}
\label{equ-3-1}
\rho_t(t,x) + \mathcal{N}[\rho(t,x)] = 0, x\in \Omega, t \in [0,T],
\end{equation}

\noindent where $ T\in \mathbb{R}^+$. Accordingly, the  continuous spatio-temporal domain $D$ is a set of points: $ D=\{(t,x)| \forall t\in [0,T], x\in [0,L] \}$. We represent this continuous domain in a discrete manner using grid points $G \in D$ that are evenly deployed throughout the domain. We define the set of grid points as $G=\{(t^{(r)},x^{(r)})|r=1,..,N_g  \}$. The total number  of grid points, $N_g$, controls the fine-grained level of $G$ as a representation of the continuous domain.

PIDL approximates $\rho(t,x)$ using a neural network with time~$t $ and location $x$ as its inputs. This neural network is called \textit{physics-uninformed neural network} (PUNN) (or \textit{estimation network} in our TSE study), which is parameterized by~$\theta$. We denote the approximation of  $\rho(t,x)$ from PUNN as $\hat{\rho}(t,x;\theta)$. During the learning phase of PUNN (i.e., to find the optimal $\theta$ for PUNN), the following residual value of the approximation $\hat{\rho}(t,x;\theta)$ is used:

\begin{equation}
\label{equ-3-2}
\hat{f}(t,x;\theta):=\hat{\rho}_t(t,x;\theta) + \mathcal{N}[\hat{\rho}(t,x;\theta)],
\end{equation}

\noindent which is designed according to the traffic flow model in Eq.~(\ref{equ-3-1}). The calculation of residual $\hat{f}(t,x;\theta)$ is done by a \textit{physics-informed neural network} (PINN). This network can compute $\hat{f}(t,x;\theta)$ directly using  $\hat{\rho}(t,x;\theta)$, the output of PUNN, as its input. When  $\hat{\rho}(t,x;\theta)$ is closer to $\rho(t,x)$, the residual will be closer to zero. PINN introduces no new parameters, and thus, shares the same $\theta$ of PUNN.

In PINN, $\hat{f}(t,x;\theta)$ is calculated by automatic differentiation technique~\cite{Baydin-2017}, which can be done by the function {\ttfamily \footnotesize tf.gradient} in Tensorflow\footnote{https://www.tensorflow.org}. The activation functions and the connecting structure of neurons in PINN are designed to conduct the differential operation in Eq.~(\ref{equ-3-2}). We would like to emphasize that, the connecting weights in PINN have fixed values which are determined by the traffic flow model, and thus, $\hat{f}$ from PINN is only parameterized by $\theta$.

The training data for PIDL consist of (1) \textit{observation points} $O=\{(t^{(i)}_o, x^{(i)}_o) | i=1,...,N_o\}$, (2) \textit{target values} $P=\{\rho^{(i)}| i=1,...,N_o\}$ (i.e., the true traffic states at the observation points), and (3) \textit{auxiliary points} $A=\{(t^{(j)}_a, x^{(j)}_a)| j=1,...,N_a\}$. $i$ and $j$ are the indexes of observation points and auxiliary points, respectively. One target value is associated with one observation point, and thus, $O$ and $P$ have the same indexing system (indexed by $i$). This paper uses the term, observed data, to denote $\{O,P\}$. Both $O$ and $A$ are subsets of the grid points $G$ (i.e., $O \in G$ and $A \in G$).  As will be seen in the future sections, if there are more equations other than Eq.~(\ref{equ-3-1}) (such as boundary conditions) that are needed to fully define the traffic flow, then more types of auxiliary points, in addition to $A$, will be created accordingly.

Observation points are usually limited to certain locations of $[0,L]$, and other areas cannot be observed. In the real world, this could represent the case in which traffic measurements over time can be made at the locations where  sensing hardware, e.g., sensors and probes, are equipped. Therefore, in general, only limited areas can be observed and we need estimation approaches to infer unobserved traffic values. In contrast to $O$, auxiliary points $A$ have neither measurement requirements nor location limitations, and the number of $A$, as well as their distributions, is controllable. Usually, $A$ are randomly distributed in the spatio-temporal domain (in implementation, $A$ is selected from $G$). As will be discussed later, auxiliary points $A$ are used for regularization purposes, making the PUNN's approximation consistent with the PDE of the traffic flow model.

To train a PUNN for TSE, the loss is defined as follows:

\vspace{-0.1in}
\begin{equation}
\label{equ-3-3}
\begin{split}
\begin{gathered}
Loss_{\theta}=\alpha \cdot MSE_o + \beta \cdot MSE_a \\
= \alpha \cdot \underbrace{\frac{1}{N_o} \sum\limits_{i=1}^{N_o} |\hat{\rho}(t^{(i)}_o, x^{(i)}_o;\theta)-\rho^{(i)}|^2}_{data\  discrepancy} \\ +  \beta \cdot \underbrace{\frac{1}{N_a}\sum\limits_{j=1}^{N_a} |\hat{f}(t^{(j)}_a, x^{(j)}_a;\theta)|^2}_{physical\  discrepancy},
\end{gathered}
\end{split}
\end{equation}

\noindent where $\alpha$ and $\beta$ are  hyperparameters for balancing the contribution  to the loss made by data discrepancy and physical discrepancy, respectively. The data discrepancy is defined as the mean square error (MSE) between approximation $\hat{\rho}$ on $O$ and target values $P$. The physical discrepancy is the MSE between residual values on $A$ and zero, quantifying the extent to which the approximation deviates from the traffic model.

Given the training data, we apply neural network training algorithms, such as backpropagation learning, to solve $\theta^* = \mathrm{argmin}_{\theta}\  Loss_{\theta}$, and this learning process is regularized by PINN via physical discrepancy. The PUNN parameterized by $\theta^*$ can then be used to approximate the traffic state at each point of $G$ (in fact, each point in the continuous domain $D$ can also be approximated). The approximation $\hat{\rho}$ is expected to be consistent with Eq.~(\ref{equ-3-1}).

\subsection{PIDL for TSE and Model Parameter Discovery}

In addition to TSE with known  PDE traffic flow, PIDL can handle traffic flow models with unknown parameters, i.e., to discover the unknown parameters $\lambda$ of the model that best describe the observed data. Let $\mathcal{N}[\ \cdot\ ;\lambda]$ be a general nonlinear differential operator parameterized by some unknown model parameters $\lambda$. Then, the problem is to  find the parameters $\lambda$ in the traffic flow model:

\begin{equation}
\label{equ-3-4}
\rho_t(t,x) + \mathcal{N}[\rho(t,x);\lambda] = 0, x\in \Omega, t \in [0,T],
\end{equation}

\noindent 
that best describe the observed data, and at the same time, approximate the traffic state $\rho(t,x)$ at each point in $G$ that satisfies this model.

For this problem, the  residual value  of traffic state approximation $\hat{\rho}(t,x;\theta)$ from PUNN is redefined as

\begin{equation}
\label{equ-3-5}
\hat{f}(t,x;\theta,\pmb{\lambda}):=\hat{\rho}_t(t,x;\theta) + \mathcal{N}[\hat{\rho}(t,x;\theta); \pmb{\lambda}].
\end{equation}

\noindent
The PINN, by which Eq.~(\ref{equ-3-5}) is calculated, is related to both $\theta$ and $\lambda$. The way in which training data are obtained and distributed remains the same as Section~\ref{3-A}. The loss function for both discovering the unknown parameters of traffic flow model and solving the TSE is defined as:

\vspace{-0.1in}
\begin{equation}
\label{equ-3-6}
\begin{split}
\begin{gathered}
Loss_{\theta,\pmb{\lambda}}=\alpha \cdot MSE_o + \beta \cdot MSE_a \\
= \alpha \cdot \underbrace{\frac{1}{N_o} \sum\limits_{i=1}^{N_o} |\hat{\rho}(t^{(i)}_o, x^{(i)}_o;\theta)-\rho^{(i)}|^2}_{data\  discrepancy} \\ +   \beta \cdot \underbrace{\frac{1}{N_a}\sum\limits_{j=1}^{N_a} |\hat{f}(t^{(j)}_a, x^{(j)}_a;\theta, \pmb{\lambda})|^2}_{physical\  discrepancy}.
\end{gathered}
\end{split}
\end{equation}

Given the training data, we apply neural network training algorithms to solve  $(\theta^*, \lambda^*) = \mathrm{argmin}_{\theta,\lambda}\  Loss_{\theta, \lambda}$. Then, the $\lambda^*$-parameterized traffic flow model of Eq.~(\ref{equ-3-4}) is the most likely physics that generates the observed data, and the $\theta^*$-parameterized PUNN can then be used to approximate the traffic states on $G$, which are consistent with the discovered traffic flow model.

\subsection{Problem Statement Summary for PIDL-based TSE}

This subsection briefly summarizes the problem statements for PIDL-based TSE. For a spatio-temporal domain represented by grid points $G=\{(t^{(r)},x^{(r)})|r=1,..,N_g  \}$, given observation points $O$, target values $P$, and auxiliary points $A$ (note $\{O,P\}$ defines the observed data)

\vspace{-0.1in}
\begin{equation}
\label{equ-3-x1}
\begin{split}
\left\{ {\begin{array}{*{20}l}
    \  O=\{(t^{(i)}_o, x^{(i)}_o) | i=1,...,N_o\}\in G \ \  \\
   \ P=\{\rho^{(i)}| i=1,...,N_o\} \  \  \\
   \ A=\{(t^{(j)}_a, x^{(j)}_a)| j=1,...,N_a\} \in G 
\end{array}} \right.
\end{split},
\end{equation}

\noindent
with the design of two neural networks: a PUNN for traffic state approximation $\hat{\rho}(t,x;\theta)$ and a PINN for computing the residual $\hat{f}(t,x;\theta)$ of $\hat{\rho}(t,x;\theta)$, then a general PIDL for TSE is to solve the problem:

\begin{equation}
\label{equ-3-x2}
\begin{split}
\begin{gathered}
\mathop {\min }\limits_\theta  Loss_{\theta}\\
where \ \ \ \ \ \ \ \ \ \ \ \ \ \ \ \ \ \ \ \ \ \ \ \ \ \ \ \ \ \ \ \ \ \ \ \ \ \ \ \ \ \ \ \ \ \ \ \ \ \ \ \   \\ Loss_{\theta} = \frac{\alpha}{N_o} \sum\limits_{i=1}^{N_o} |\hat{\rho}(t^{(i)}_o, x^{(i)}_o;\theta)-\rho^{(i)}|^2 \\+  \frac{\beta}{N_a}\sum\limits_{j=1}^{N_a} |\hat{f}(t^{(j)}_a, x^{(j)}_a;\theta)|^2 ,\\
\hat{f}(t,x;\theta):=\hat{\rho}_t(t,x;\theta) + \mathcal{N}[\hat{\rho}(t,x;\theta)].
\end{gathered}
\end{split}
\end{equation}

\noindent
The PUNN parameterized by the solution $\theta^{*}$ can then be used to approximate the traffic states on $G$. 

In contrast, a general  PIDL for both TSE and model parameter discovery is to solve the problem:

\begin{equation}
\label{equ-3-x3}
\begin{split}
\begin{gathered}
\mathop {\min }\limits_{\theta,\lambda} Loss_{\theta,\lambda}\\
where \ \ \ \ \ \ \ \ \ \ \ \ \ \ \ \ \ \ \ \ \ \ \ \ \ \ \ \ \ \ \ \ \ \ \ \ \ \ \ \ \ \ \ \ \ \ \ \ \ \ \ \   \\ Loss_{\theta,\lambda} = \frac{\alpha}{N_o} \sum\limits_{i=1}^{N_o} |\hat{\rho}(t^{(i)}_o, x^{(i)}_o;\theta)-\rho^{(i)}|^2  \\ +   \frac{\beta}{N_a}\sum\limits_{j=1}^{N_a} |\hat{f}(t^{(j)}_a, x^{(j)}_a;\theta, \lambda)|^2 ,\\
\hat{f}(t,x;\theta,\lambda):=\hat{\rho}_t(t,x;\theta) + \mathcal{N}[\hat{\rho}(t,x;\theta);\lambda],
\end{gathered}
\end{split}
\end{equation}

\noindent
where residual $\hat{f}$ is related to both PUNN parameters $\theta$ and traffic flow model parameters $\lambda$. The PUNN parameterized by  the solution $\theta^{*}$ can then be used to approximate the traffic states on $G$, and  solution $\lambda^{*}$ is the most likely model parameters that describe the observed data.

In the next three sections, we will demonstrate PIDL's ability to estimate  traffic density dynamics and discover model parameters using two popular highway traffic flow models, Greenshields-based LWR and three-parameter-based LWR. Then, we extend the PIDL-based approach to a real-world highway  scenario in the NGSIM data.

\section{PIDL for Greenshields-Based LWR}
\label{sec-IV}

This example aims to show the ability of our method to estimate the traffic dynamics governed by the LWR model based on Greenshields flux function.

Define flow rate $Q$ to be the number of vehicles passing a specific position on the road per unit time, and traffic density $\rho$ to be the average number of vehicles per unit length of the road. By defining $u$ as the average speed of a specific position on the road, we can deduce $Q=\rho u$. The traffic flux $Q(\rho)$ describes  $Q$ as a function of  $\rho$. We treat $\rho$ as the basic traffic state variable to estimate. Greenshields flux~\cite{Greenshields-1935} is a basic and popular choice of flux function, which is defined as $Q(\rho)=\rho u_{max}(1-\rho/\rho_{max})$, where $u_{max}$ and $\rho_{max}$ are maximum velocity and maximum (jam) density, respectively. This flux function has a quadratic form with two coefficients $u_{max}$ and $\rho_{max}$, which are usually fitted with data.

The LWR model~\cite{Lighthill-1955, Richards-1956} describes the macroscopic traffic flow dynamics as $\rho_t+(Q(\rho))_x=0$, which is derived from a conservation law of vehicles. In order to reproduce more phenomena in observed traffic data, such as delayed driving behaviors due to drivers' reaction time, diffusively corrected LWRs were introduced, by adding a diffusion term, containing a second-order derivative $\rho_{xx}$~\cite{Nelson-2002, LiJia-2010, Burger-2013, Acosta-2015, Burger-2017}. We focus on one version of the diffusively corrected LWRs:  $\rho_t+(Q(\rho))_x=\epsilon \rho_{xx}$, where $\epsilon$ is the diffusion coefficient.

In this section, we study the Greenshields-based LWR traffic flow model of a ``ring road":

\begin{equation}
\label{equ-3-7}
\begin{split}
\begin{gathered}
\rho_t + (Q(\rho))_x=\epsilon \rho_{xx}, \  t\in [0,3], \  x\in [0,1],\\
Q(\rho)=\rho \cdot u_{max}\Bigl( 1- \frac{\rho}{\rho_{max}} \Bigl), \\
\rho(t,0)=\rho(t,1) \ \ \ (boundary\ condition\ 1), \\
\rho_x(t,0)=\rho_x(t,1) \ \ \ (boundary\ condition\ 2),
\end{gathered}
\end{split}
\end{equation}

\noindent
where $\rho_{max}=1.0$, $u_{max}=1.0$, and $\epsilon = 0.005 $. $\rho_{max}$ and $u_{max}$ are usually determined by physical restrictions of the road and vehicles.

\begin{figure}[h!]
\centering
  \includegraphics[scale=0.7]{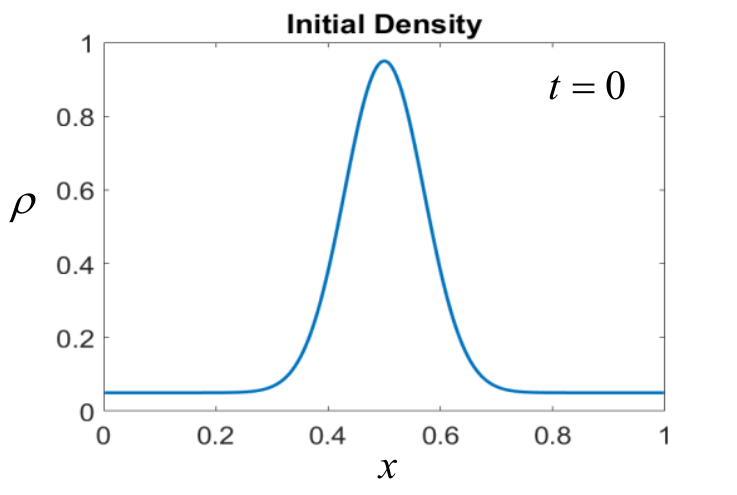}
  \caption{Bell-shape initial traffic density $\rho$ over $x\in [0,1]$ at $t=0$.}
  \label{fig:ch4-initial}
\end{figure}

\begin{figure}[h!]
\centering
  \includegraphics[scale=0.61]{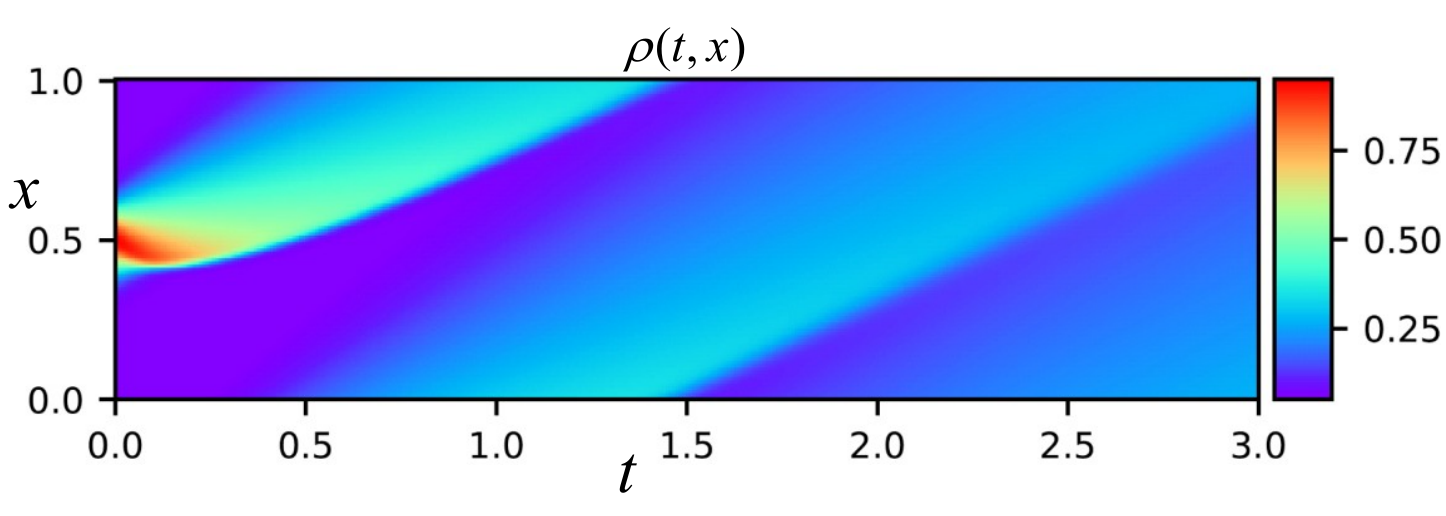}
  \caption{Numerical solution of Eqs.~(\ref{equ-3-7}) using the Godunov scheme. We treat the numerical solution as the ground truth in our TSE experiments.}
  \label{fig:ch4-exact}
\end{figure}

Given the bell-shape initial showed in Fig.~\ref{fig:ch4-initial}, we apply the Godunov scheme~\cite{Godunov-1959} to solve Eqs.~(\ref{equ-3-7}) on 960 (time) $\times$ 240 (space) grid points $G$ evenly deployed throughout the $[0,3]\times [0,1]$  domain. In this case, the total number of grid points $G$ is $N_g=$960$\times$240. The numerical solution is shown in Fig.~\ref{fig:ch4-exact}. From the figure, we can visualize the dynamics as follows: At $t=0$, there is a peak density at the center of the road segment, and this peak evolves to propagate along the direction of $x$, which is known as the phenomenon of  traffic shockwave. Since this is a ring road, the shockwave reaching $x=1$ continues at $x=0$. This numerical solution of the Greenshields-based LWR model is treated as the ground-truth traffic density. We will apply a PIDL-based approach method to estimate the entire traffic density field using observed data.

\subsection{PIDL Architecture Design}

Based on Eqs.~(\ref{equ-3-7}), we define the residual value of PUNN's traffic density estimation $\hat{\rho}(t,x;\theta)$ as

\begin{equation}
\label{equ-8}
\hat{f}(t,x;\theta) := \hat{\rho}_t(t,x;\theta) + (Q(\hat{\rho}(t,x;\theta)))_x-\epsilon \hat{\rho}_{xx}(t,x;\theta).
\end{equation}

\noindent The residual value is calculated by a PINN. 

\begin{figure}[t!]
\centering
  \includegraphics[scale=0.71]{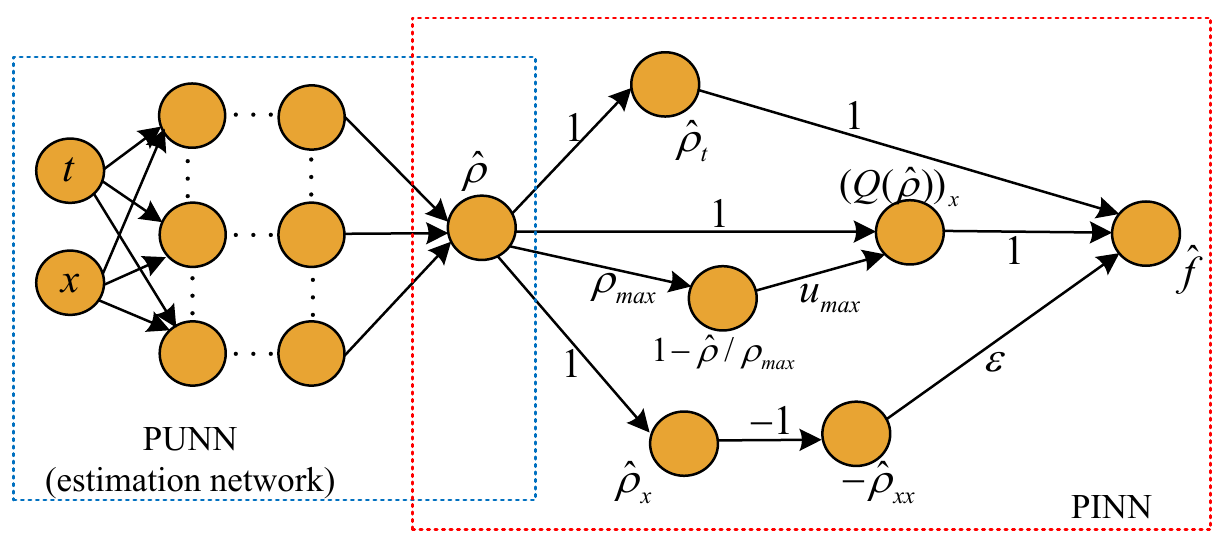}
  \caption{PIDL architecture for Greenshields-based LWR in Eqs.~(\ref{equ-3-7}), consisting of a PUNN for traffic density estimation and a PINN for calculating the residual Eq.~(\ref{equ-8}). For experiments of estimation without  parameter discovery, all connecting weights, including $\rho_{max}=1$, $u_{max}=1$, and $\epsilon = 0.005$, are known and fixed in PINN. For experiments of estimation with parameter discovery, $\rho_{max}$, $u_{max}$, and $\epsilon$ are learning variables in PINN.}
  \label{fig:ch4-PINN_structure}
\end{figure}

Given the definition of $\hat{f}$, the corresponding PIDL architecture that encodes Greenshields-based LWR model is shown in Fig.~\ref{fig:ch4-PINN_structure}. This architecture consists of a PUNN for traffic density estimation, followed by a PINN for calculating the residual Eq.~(\ref{equ-8}). PUNN parameterized by $\theta$ is designed as a fully-connected feedforward neural network with 8 hidden layers and 20  hidden nodes in each hidden layer. Hyperbolic tangent function (tanh)  is used as the activation function for each hidden neuron in PUNN. In contrast, in PINN, connecting weights are fixed and the activation function of each node is designed to conduct specific nonlinear operation for calculating an intermediate value of~$\hat{f}$.

To customize the training of PIDL to  Eqs.~(\ref{equ-3-7}), in addition to the training data $O$, $P$ and $A$ defined in Section~\ref{3-A}, we  need to introduce \textit{ boundary auxiliary points} $B=\{(t^{(k)}_b,0)| k = 1,...,N_b\} \cup \{(t^{(k)}_b,1)| k = 1,...,N_b\}$, for learning the two boundary conditions in Eqs.~(\ref{equ-3-7}).

For experiments of state estimation without model parameter discovery, where the PDE parameters are known, we design the following loss:

\vspace{-0.1in}
\begin{equation}
\label{equ-9}
\begin{split}
\begin{gathered}
Loss_{\theta}=\alpha \cdot MSE_o +  \beta \cdot MSE_a + \gamma \cdot MSE_{b1} + \eta \cdot MSE_{b2} \\
= \frac{\alpha}{N_o} \sum\limits_{i=1}^{N_o} |\hat{\rho}(t^{(i)}_o, x^{(i)}_o;\theta)-\rho^{(i)}|^2 +    \frac{\beta}{N_a}\sum\limits_{j=1}^{N_a} |\hat{f}(t^{(j)}_a, x^{(j)}_a;\theta)|^2 \\
+ \frac{\gamma}{N_b} \sum\limits_{k=1}^{N_b} |\hat{\rho}(t^{(k)}_b, 0;\theta)-\hat{\rho}(t^{(k)}_b, 1;\theta)|^2 \\
+ \frac{\eta}{N_b} \sum\limits_{k=1}^{N_b} |\hat{\rho}_x(t^{(k)}_b, 0;\theta)-\hat{\rho}_x(t^{(k)}_b, 1;\theta)|^2 ,
\end{gathered}
\end{split}
\end{equation}

\noindent where each value used by the loss is an output of certain node of the PINN (see Fig.~\ref{fig:ch4-PINN_structure}). $MSE_{b1}$, scaled by $\gamma$, is the mean square error between  estimations at the two boundaries $x=0$ and $x=1$. $MSE_{b2}$, scaled by $\eta$, quantifies the difference of first order derivatives at the two boundaries.

\subsection{TSE using Initial Observation}
\label{IV-B}

\begin{figure}[h!]
\centering
  \includegraphics[scale=0.61]{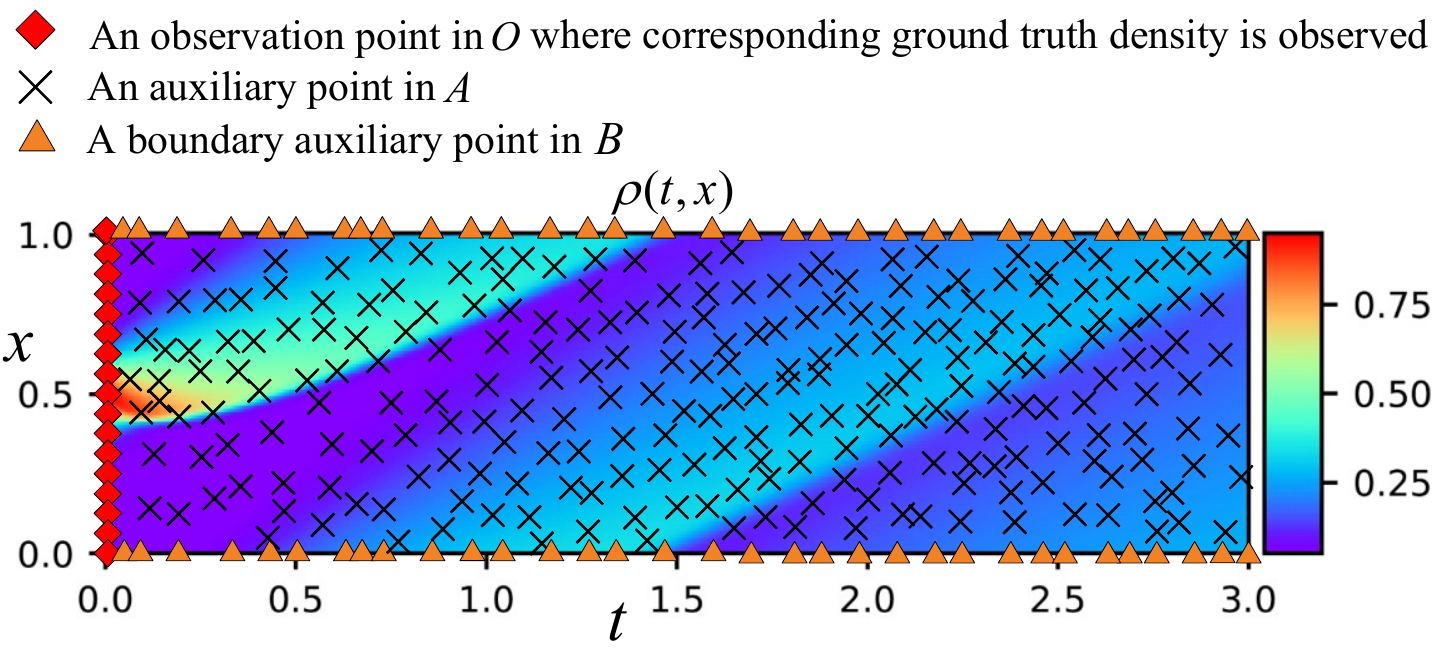}
  \caption{A sparse presentation of the deployment of observation points $O$ at the initial $t=0$, auxiliary points $A$ randomly selected from $G$, and boundary auxiliary points $B$ deployed  at the boundaries $x=0$ and $x=1$ for certain time points.}
  \label{fig:deployment}
\end{figure}

We start with justifying the ability of PIDL in Fig.~\ref{fig:ch4-PINN_structure} for estimating the traffic density field in Fig.~\ref{fig:ch4-exact} using the observation of the entire road at  $t=0$ (i.e., the initial traffic density condition). To be specific, 240 grid points along the space dimension at $t=0$ are used as the observation points $O$ with $N_o=240$, and their corresponding densities constitute the target values $P$ for training. There are $N_a= 100,000$ auxiliary points in $A$ randomly selected from grid points $G$. $N_b=650$ out of 960 \textit{grid time points} (i.e., the time points on the temporal dimension of $G$) are randomly selected to create boundary  auxiliary  points $B$. A sparse version of the deployment of $O$, $A$ and $B$ in the spatio-temporal domain is shown in Fig.~\ref{fig:deployment}. Each observation point is associated with a target value in $P$. Note $O$, $A$ and $B$ are all subsets of $G$.

We train the proposed PIDL on an NVIDIA Titan RTX GPU with 24 GB memory. By default, we use the $\mathbbm{L}^2$ relative error on $G$ to quantify the estimation error of the entire domain:

\begin{equation}
\label{equ-10}
Err= \frac{\sqrt{\sum_{r=1}^{N_g} \bigl|\hat{\rho}(t^{(r)}, x^{(r)};\theta)-\rho(t^{(r)}, x^{(r)}) \bigl|^2}}{\sqrt{\sum_{r=1}^{N_g} \bigl|\rho(t^{(r)}, x^{(r)}) \bigl|^2}}.
\end{equation}

\noindent
The reason for choosing the $\mathbbm{L}^2$ relative error is to normalize the estimation inaccuracy, mitigating the influence from the scale of true density values.

\begin{figure}[h!]
\centering
  \includegraphics[scale=0.61]{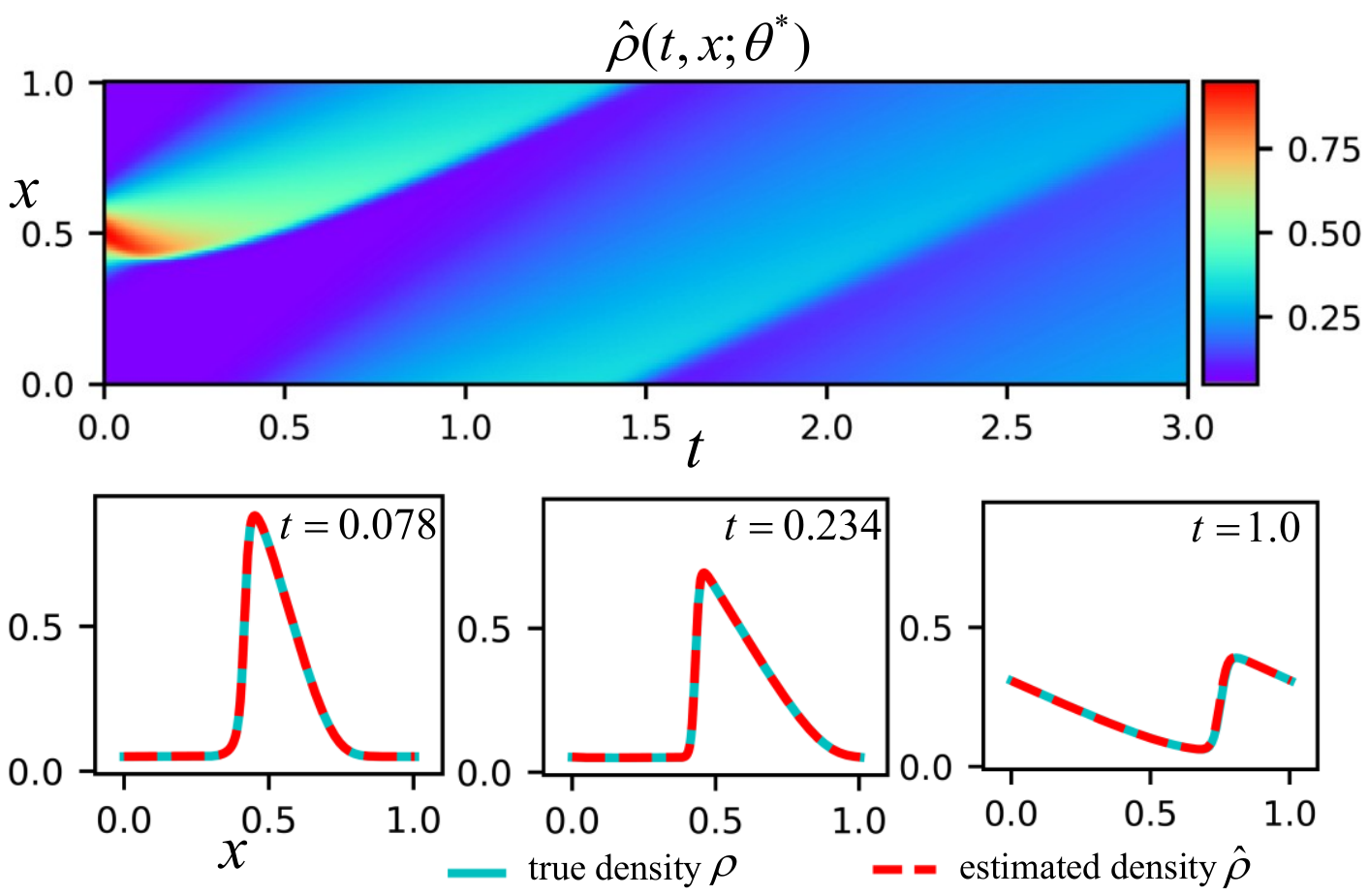}
  \caption{Top: Estimation of the traffic density dynamics $\hat{\rho}(t,x;\theta^*)$ on  grid points $G$ in the domain using the trained PUNN. Bottom: Snapshots  of  estimated and true traffic density at certain time points.}
  \label{fig:ch4-results}
\end{figure}

We use the Xavier uniform initializer~\cite{Xavier-2010} to initialize $\theta$ of PUNN. This neural network initialization method takes the number of a layer's incoming and outgoing network connections into account when initializing the weights of that layer, which may lead to a good convergence. Then, we train the PUNN through the PIDL architecture using a popular stochastic gradient descent algorithm, the Adam optimizer~\cite{Konur-2015}, for $2,000$ steps for a rough training. A follow-up fine-grained training is done by the L-BFGS optimizer~\cite{Byrd-1995} for stabilizing the convergence, and the process terminates until the loss change of two consecutive steps is no larger than $10^{-16}$. This training process converges to a local optimum $\theta^*$ that minimizes the loss in Eq.~(\ref{equ-9}).

The results of applying the PIDL to Greenshields-based LWR is presented in Fig.~\ref{fig:ch4-results}, where PUNN is parameterized by the optimal $\theta^*$. As shown in Fig.~\ref{fig:ch4-results}, the estimation $\hat{\rho}(t,x;\theta^*)$ is visually the same as the true dynamics $\rho(t,x)$ in Fig.~\ref{fig:ch4-exact}. By looking into the estimated and true traffic density over $x$ at certain time points, there is a good agreement between two traffic density curves. The $\mathbbm{L}^2$ relative error of the  estimation to the true traffic density is $1.6472\times 10^{-2}$. Empirically, the difference cannot be visually distinguished when the estimation error is smaller than $6\times 10^{-2}$.

\subsection{TSE using Observation from Loop Detectors}

We now apply PIDL to the same TSE problem, but using observation from loop detectors, i.e., only the traffic density at certain locations where loop detectors are installed can be observed. By default, loop detectors are evenly located along the road. \textcolor{black}{We would like to clarify that in this paper, the training data are the observed data from detectors, i.e., the traffic states on the points at certain locations where loops are equipped. As implied by Eq.~(\ref{equ-10}), the test data are the traffic states on the grid points $G$, which represent the whole spatio-temporal domain of the selected highway segment (i.e., a fixed region). The test data consist of the observed data (i.e., the training data) and the data that are not observed (the test data is defined in this way per the definition of TSE used in this paper).}

\begin{figure}[h!]
\centering
  \includegraphics[scale=0.51]{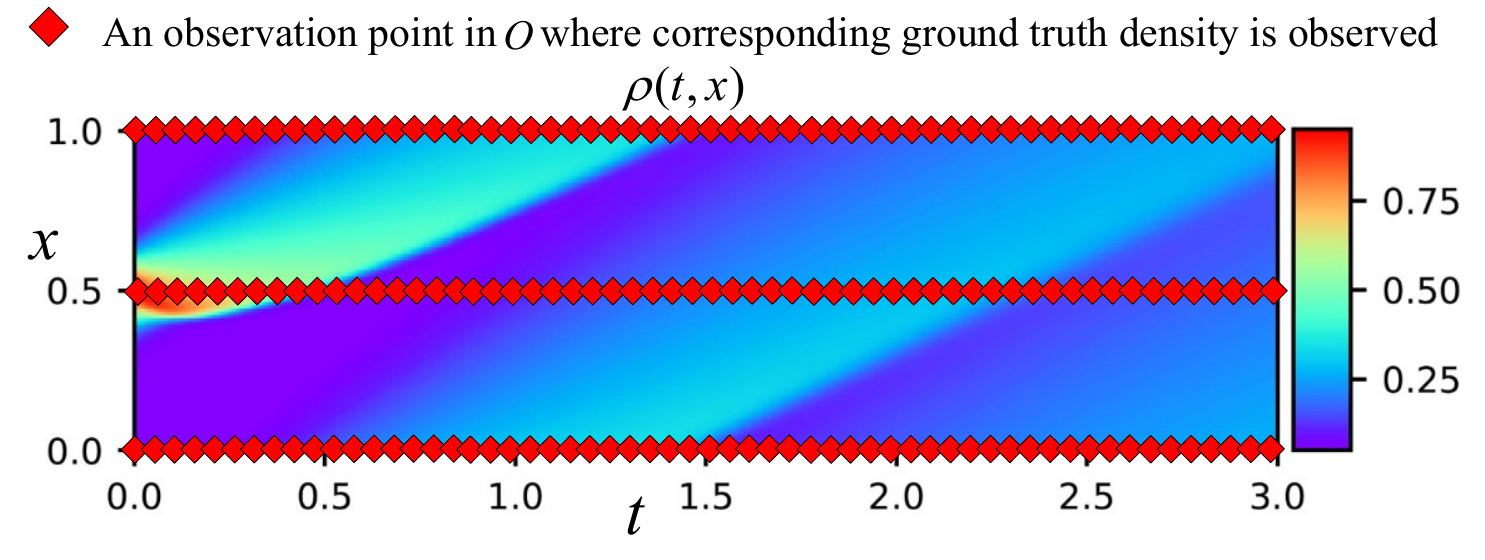}
  \caption{A sparse presentation of the deployment of $N_o=3\times 960$ observation points $O$ for three loop detectors. Corresponding density target values $P$ are collected by the loop detectors. The deployment of auxiliary points $A$ and boundary auxiliary points $B$ remains the same as Fig.~\ref{fig:deployment}. }
  \label{fig:ch4-3loops}
\end{figure}

Fig.~\ref{fig:ch4-3loops} shows the deployment of observation points $O$ for three loop detectors, where the detectors locate evenly on the road at $x=0$, $x=0.5$ and $x=1$. The traffic density records of those locations over time constitute the  target values $P$.

\begin{figure}[h!]
\centering
  \includegraphics[scale=0.65]{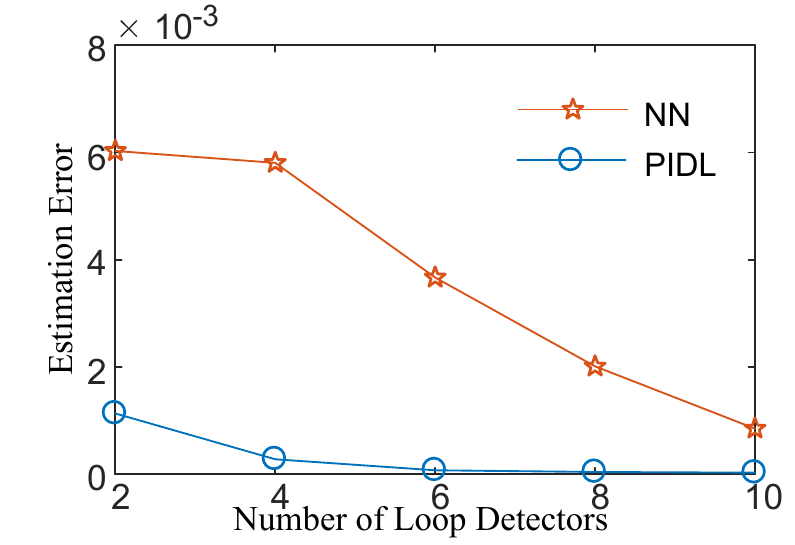}
  \caption{Greenshields-based PIDL estimation error vs NN over different loop detectors.}
  \label{fig:ch4-infer1}
\end{figure}

Using the same training process in Section~\ref{IV-B}, we conduct PIDL-based TSE experiments with different numbers of loop detectors. For each number of loop detectors, we make 100 trails with hyperparamters randomly selected between $0$ and $200$. 
\textcolor{black}{The minimal-achievable estimation errors of PIDL over the numbers of loop detectors are presented in Fig.~\ref{fig:ch4-infer1}, which are compared with the pure NN that is not regularized by the physics. The estimation errors of both methods decrease as more loop detectors are installed to collect  data, and the PIDL outperforms the NN especially when the loop numbers are small.}
The results  demonstrate that  the PIDL-based approach is data efficient as it can handle the TSE task  with  fewer loop  detectors  for  the traffic dynamics governed by the Greenshields-based LWR.

\subsection{TSE and Parameter Discovery using  Loop Detectors}

This subsection justifies  the  ability  of  the  PIDL architecture in Fig.~\ref{fig:ch4-PINN_structure} for dealing with both TSE and model parameter discovery. In this experiment, three model parameters $\rho_{max}$, $u_{max}$ and $\epsilon$ are encoded as learning variables in PINN. We denote $\lambda=(\rho_{max},u_{max},\epsilon)$ in this experiment. The residual and loss become

\vspace{-0.1in}
\begin{equation}
\label{equ-11}
\begin{split}
 \hat{f}(t,x;\theta,\pmb{\lambda})  := & \hat{\rho}_t(t,x;\theta)    +(Q(\hat{\rho}(t,x;\theta);\pmb{\rho_{max}},\pmb{u_{max}}))_x \\
 & -\pmb{\epsilon} \hat{\rho}_{xx}(t,x;\theta) ,
\end{split}
\end{equation}

\noindent and

\vspace{-0.1in}
\begin{equation}
\label{equ-12}
\begin{split}
Loss_{\theta,\pmb{\lambda}} &=  \frac{\alpha}{N_o} \sum\limits_{i=1}^{N_o} |\hat{\rho}(t^{(i)}_o, x^{(i)}_o;\theta)-\rho^{(i)}|^2 \\
&  +    \frac{\beta}{N_a}\sum\limits_{j=1}^{N_a} |\hat{f}(t^{(j)}_a, x^{(j)}_a;\theta,\pmb{\lambda})|^2 \\
&+ \frac{\gamma}{N_b} \sum\limits_{k=1}^{N_b} |\hat{\rho}(t^{(k)}_b, 0;\theta)-\hat{\rho}(t^{(k)}_b, 1;\theta)|^2 \\
&+ \frac{\eta}{N_b} \sum\limits_{k=1}^{N_b} |\hat{\rho}_x(t^{(k)}_b, 0;\theta)-\hat{\rho}_x(t^{(k)}_b, 1;\theta)|^2 ,
\end{split}
\end{equation}

\noindent respectively. We use $N_a = 150,000$ auxiliary points and other experimental configurations remain unchanged. We conduct PIDL-based TSE experiments using different numbers of loop detectors to solve $(\theta^*, \lambda^*) = \mathrm{argmin}_{\theta,\lambda}\  Loss_{\theta, \lambda}$. In addition to the traffic density estimation errors of $\hat{\rho}(t,x;\theta^*)$, we evaluate the estimated model parameters $\lambda^*$ using the $\mathbbm{L}^2$ relative error and present them in percentage. The results are shown in Table~\ref{tab:ch4}, where the errors below the dashed line are acceptable.

\begin{table}[h!]
    \caption{Errors on Estimated Traffic Density and Model Parameters}
    \label{tab:ch4}
    \centering
    \begin{threeparttable}
    \begin{tabular}{ccccc}
    \toprule
    \toprule
    \multicolumn{1}{c}{$m$} & \multicolumn{1}{c}{$\hat{\rho}(t,x;\theta^*)$} & \multicolumn{1}{c}{$\rho^*_{max}$(\%)} & \multicolumn{1}{c}{$u^*_{max}$(\%)} &
    \multicolumn{1}{c}{$\epsilon^*$(\%)}\\
    \midrule
    2     & 6.007$\times 10^{-1}$ & 982.22 & 362.50 & $>$1000\\ \hdashline
    3     & 4.878$\times 10^{-2}$ & 0.53 & 0.05 & 1.00\\
    4     & 3.951$\times 10^{-2}$ & 0.95 & 0.54 & 4.37\\
    5     & 3.881$\times 10^{-2}$ & 0.16 & 0.14 & 4.98\\
    6     & 2.724$\times 10^{-2}$ & 0.07 & 0.18 & 1.38\\
    8     & 3.441$\times 10^{-3}$ & 0.25 & 0.47 & 1.40\\
    \bottomrule
    \bottomrule
    \end{tabular}
    \begin{tablenotes}
      \footnotesize
      \item $m$ stands for the number of loop detectors.  $\lambda^*=(\rho^*_{max},u^*_{max},\epsilon^*)$ are estimated parameters, compared to the true parameters $\rho_{max}=1,u_{max}=1,\epsilon=0.005$.
    \end{tablenotes}
  \end{threeparttable}
\end{table}

From the table, we can see that the traffic density estimation errors improve as the number of loop detectors increases. When more than two loop detectors are used, the learning model parameters are able to converge to the true parameters $\lambda$. Specifically, for three loop detectors, in addition to a good traffic density estimation error of 4.878$\times 10^{-2}$, the model parameters converge to $\rho^*_{max}=1.00532$, $u^*_{max}=0.99955$ and $\epsilon^*=0.00495$, which are very close to the true ones. The results demonstrate that PIDL method can handle both TSE and model parameter discovery with three loop detectors for the traffic dynamics from the Greenshields-based LWR.

\section{PIDL for Three-Parameter-Based LWR}
\label{sec-V}

This example aims to show the ability of our method to handle the traffic dynamics governed by the three-parameter-based LWR.

The traffic flow model considered in this example is the same as Eqs.~(\ref{equ-3-7}) except for a different flux function $Q(\rho)$: three-parameter flux function~\cite{Fan-2014,Fan-2013}. This flux function is triangle-shaped, which is a differentiable version of the piecewise linear Daganzo-Newell flux function~\cite{Daganzo-1994,Newell-1993}. This function is defined as:

\vspace{-0.1in}
\begin{equation}
\label{equ-13}
\begin{split}
 Q(\rho)&=\sigma \Bigl(a+(b-a)\frac{\rho}{\rho_{max}}-\sqrt{1+y^2}\Bigl),\\
 &a=\sqrt{1+\bigl(\delta p\bigl)^2},\\
 &b=\sqrt{1+\bigl(\delta (1-p)\bigl)^2},\\
 & y = \delta \Bigl(\frac{\rho}{\rho_{max}}-p  \Bigl),
\end{split}
\end{equation}

\noindent where  $\delta$, $p$ and $\sigma$ are the three free parameters as the function is named. The parameters $\sigma$ and $p$ control the maximum flow rate and critical density (where the flow is maximized), respectively. $\delta$ controls the roundness level of $Q(\rho)$. In addition to the above-mentioned three parameters, $\rho_{max}$ and diffusion coefficient $\epsilon$ are also part of the model parameters.

\begin{figure}[h!]
\centering
  \includegraphics[scale=0.61]{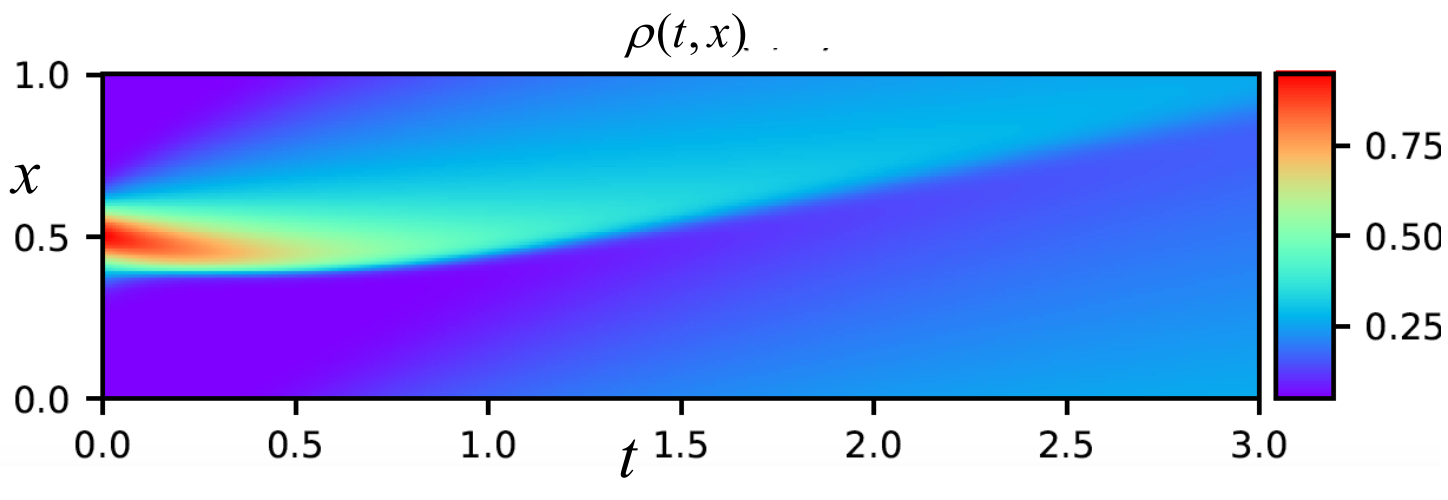}
  \caption{Numerical solution of three-parameter-based LWR on grid points $G$ using the Godunov scheme. The numerical solution serves as the ground truth.}
  \label{fig:ch5-exact}
\end{figure}

Using the same domain and grid points setting in Section~\ref{sec-IV} and initializing the dynamics with the bell-shaped density in Fig.~\ref{fig:ch4-initial}, the numerical solution of three-parameter-based LWR is presented in Fig.~\ref{fig:ch5-exact}. The parameters are given as $\delta = 5$, $p=2$, $\sigma = 1$, $\rho_{max}=1$ and $\epsilon = 0.005$. We treat the numerical solution as the ground truth and conduct our TSE experiments.

\subsection{PIDL Architecture Design}

\begin{figure}[h!]
\centering
  \includegraphics[scale=0.62]{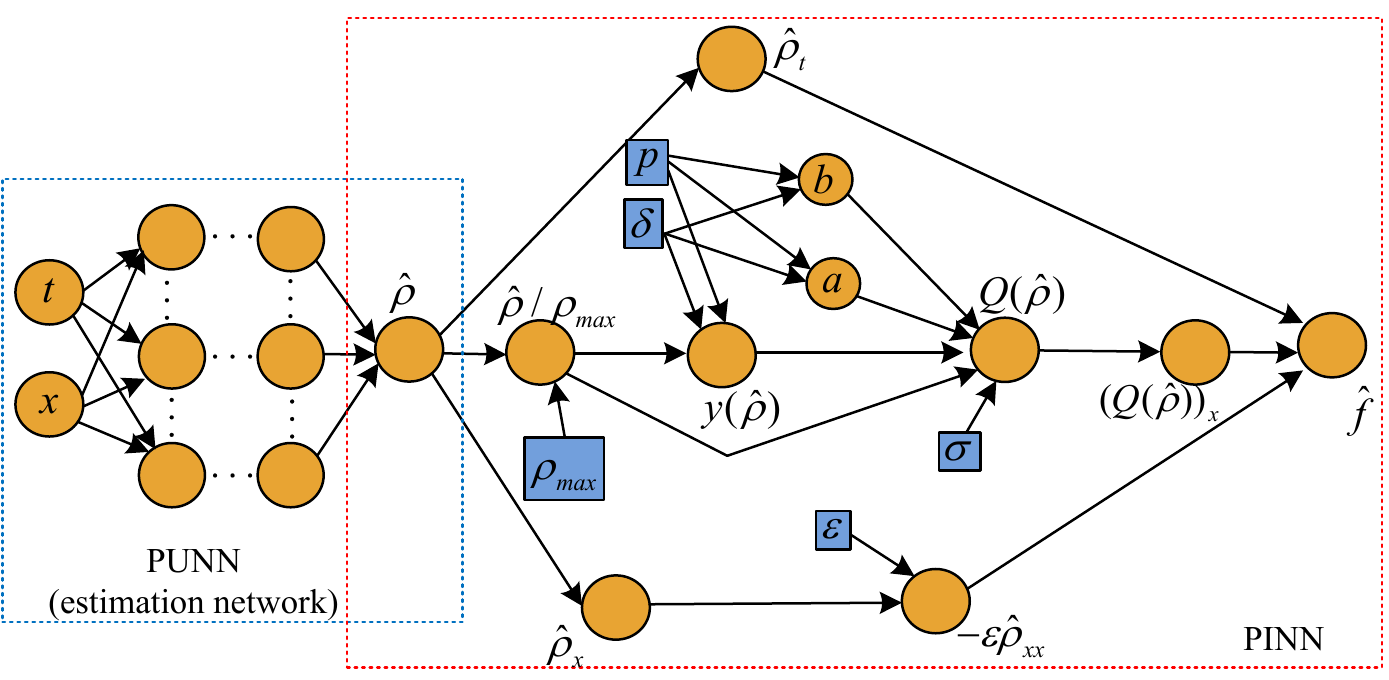}
  \caption{PIDL architecture for three-parameter-based LWR using flux $Q(\rho)$ defined in Eq.~(\ref{equ-13}). Model parameters are held by variable nodes (blue rectangular nodes). Connecting weights in PINN are one by default and specific operations for intermediate values are calculated by  PINN nodes via their activation functions. For experiments of  estimation without  parameter discovery, model parameters $\delta = 5$, $p=2$, $\sigma = 1$, $\rho_{max}=1$ and $\epsilon = 0.005$, are known and fixed in PINN. For experiments of estimation with parameter discovery, model parameters are learning variables in PINN.}
  \label{fig:ch5-PINN}
\end{figure}

Using the definition of the residual value $\hat{f}$ in Eq.~(\ref{equ-8}), the corresponding PIDL architecture that encodes three-parameter-based LWR is shown in Fig.~\ref{fig:ch5-PINN}. Different from Fig.~\ref{fig:ch4-PINN_structure} where model parameters can be easily encoded as connecting weights, there are too many intermediate values involving the same model parameter, and thus, we cannot simply encode the model parameters as connecting weights. Instead, we use variable nodes (see the blue rectangular nodes in the graph) for holding model parameters, such that we can link the model parameter nodes to neurons that are directly related.

For the training loss, Eq.~(\ref{equ-9}) is used, which includes the regularization of minimizing the residual and the difference between the density values at the two boundaries.

\subsection{TSE using Initial Observation}

We justify the ability of the PIDL architecture in Fig.~\ref{fig:ch5-PINN} for estimation of the traffic density field in Fig.~\ref{fig:ch5-exact} using the observation of the initial traffic density of the entire road at $t=0$. The training data selection is the same as Section~\ref{IV-B}, except for that $N_a=150,000$ auxiliary points $A$ are used, because this is a relatively more complicated model.

\begin{figure}[h!]
\centering
  \includegraphics[scale=0.61]{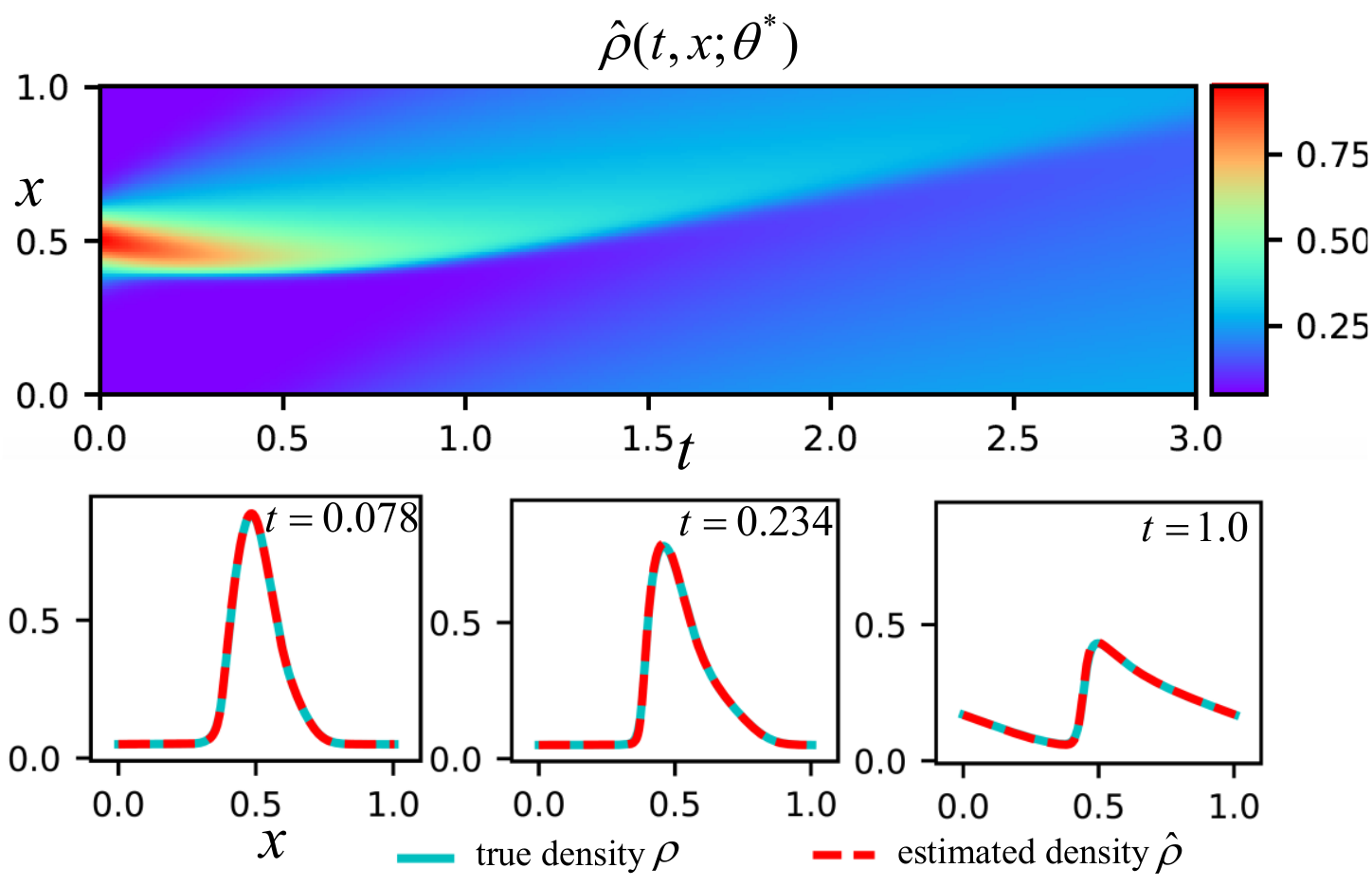}
  \caption{Top: Estimation of the traffic density dynamics $\hat{\rho}(t,x;\theta^*)$ on $G$ in the domain using the trained PUNN for three-parameter-based LWR. Bottom: Snapshots  of  estimated and true traffic density at certain time points.}
  \label{fig:ch5-results}
\end{figure}

The training process finds the optimal $\theta^*$ for PUNN, that minimizes the loss in Eq.~(\ref{equ-9}). The results of applying the $\theta^*$-parameterized PUNN to the three-parameter-based LWR dynamics estimation are presented in Fig.~\ref{fig:ch5-results}. The performance of the trained PUNN is  satisfactory as the estimation $\hat{\rho}(t,x;\theta^*)$ achieves an $\mathbbm{L}^2$ relative error of $1.0347\times 10^{-2}$, which is much smaller than the distinguishable error  $6\times 10^{-2}$.

\subsection{TSE using Observation from Loop Detectors}
\label{V-C}

\begin{figure}[h!]
\centering
  \includegraphics[scale=0.65]{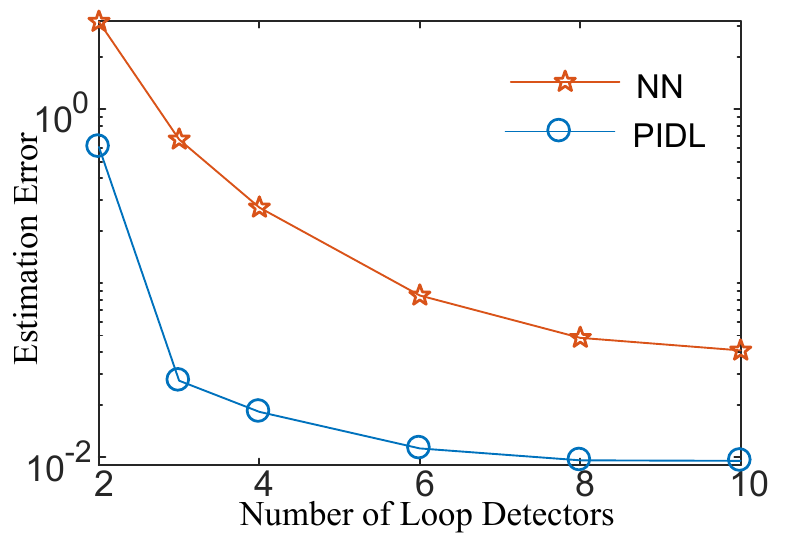}
  \caption{3 parameter-based PIDL estimation error vs NN over different loop detectors.}
  \label{fig:ch5-infer1}
\end{figure}

\textcolor{black}{We conduct PIDL-based TSE experiments with different numbers of loop detectors and compare the results with NN. Fig.~\ref{fig:ch5-infer1} shows the PIDL approach achieves better errors than the pure NN, justifying the benefits of PIDL.}
A logarithmic error-axis is used for better visualizing the difference in estimation errors.

From the figure, we observe that the estimation errors decrease as more loop detectors are installed to collect  data. A significant improvement in estimation quality is achieved when three loop detectors are installed, obtaining an error of $2.745 \times 10^{-2}$ compared to $6.066 \times 10^{-2}$ for two loop detectors. The NN needs more detectors to achieve these low errors.

\subsection{TSE and Parameter Discovery using Loop Detectors}

This experiment focuses on using PIDL to address both TSE and model parameter discovery. Five learning variables $\lambda = (\delta,p,\sigma,\rho_{max},\epsilon)$ need to be determined. The residual becomes:

\vspace{-0.1in}
\begin{equation}
\label{equ-14}
\begin{split}
 \hat{f}(t,x;\theta,\pmb{\lambda})  & :=  \hat{\rho}_t(t,x;\theta) \\ + & (Q(\hat{\rho}(t,x;\theta);\pmb{\delta,p,\sigma, \rho_{max}}))_x 
  -\pmb{\epsilon} \hat{\rho}_{xx}(t,x;\theta) ,
\end{split}
\end{equation}

\noindent
and the $Loss_{\theta,\pmb{\lambda}}$ in Eq.~(\ref{equ-12}) is used in the training process. All experimental configurations and training process remain the same as Section~\ref{V-C}. We solve $(\theta^*, \lambda^*) = \mathrm{argmin}_{\theta,\lambda}\  Loss_{\theta, \lambda}$, and the results of traffic density estimation and model parameter discovery are presented in Table~\ref{tab:ch5}. The errors below the dashed line are acceptable.

\begin{table}[h!]
    \caption{Errors on Estimated Traffic Density and Model Parameters}
    \label{tab:ch5}
    \centering
    \begin{threeparttable}
    \begin{tabular}{ccccccc}
    \toprule
    \toprule
    \multicolumn{1}{c}{$m$} & 
    \multicolumn{1}{c}{$\hat{\rho}(t,x;\theta^*)$} & 
    \multicolumn{1}{c}{$\delta^*$(\%)} & 
    \multicolumn{1}{c}{$p^*$(\%)} &
    \multicolumn{1}{c}{$\sigma^*$(\%)} &
    \multicolumn{1}{c}{$\rho^*_{max}$(\%)} &
    \multicolumn{1}{c}{$\epsilon^*$(\%)}\\
    \midrule
    2     & 1.2040$\times 10^{0}$ & 58.19 & 153.72 & $>$1000 & $>$1000 & 99.98\\ 
    3     & 7.550$\times 10^{-1}$ & 54.15 & 124.52 & $>$1000 & $>$1000 & 99.95\\
    4     & 1.004$\times 10^{-1}$ & 59.07 & 72.63 & 381.31 & 14.60 & 6.72\\
    \hdashline
    5     & 3.186$\times 10^{-2}$ & 2.75 & 4.03 & 6.97 & 0.29 & 3.00\\
    6     & 1.125$\times 10^{-2}$ & 0.69 & 2.49 & 2.26 & 0.49 & 7.56\\
    8     & 7.619$\times 10^{-3}$ & 1.03 & 2.43 & 3.60 & 0.30 & 7.85\\
    12     & 5.975$\times 10^{-3}$ & 1.83 & 1.65 & 1.24 & 0.82 & 5.70\\
    \bottomrule
    \bottomrule
    \end{tabular}
    \begin{tablenotes}
      \footnotesize
      \item $m$ stands for the number of loop detectors.  $\lambda^*=(\delta^*,p^*,\sigma^*,\rho^*_{max},\epsilon^*)$ are estimated parameters, compared to the true parameters $\delta=5, p=2, \sigma=1, \rho_{max}=1, \epsilon=0.005$.
    \end{tablenotes}
  \end{threeparttable}
\end{table}

From the table, we observe that  the PIDL architecture in Fig.~\ref{fig:ch5-PINN} with five loop detectors can achieve a satisfactory performance on both traffic density estimation and model parameter discovery. In general, more loop detectors can help our model improve the TSE accuracy, as well as the convergence to the true model parameters. Specifically, for five loop detectors, an estimation error of $3.186 \times 10^{-2}$ is obtained, and the model parameters converge to $\delta^* = 4.86236$, $p^* = 0.19193$, $\sigma^*=0.10697$, $\rho^*_{max}=1.00295$ and $\epsilon^*=0.00515$, which are decently close to the ground truth. The results demonstrate that PIDL can handle both TSE and model parameter discovery with five loop detectors for the traffic dynamics governed by the  three-parameter-based LWR.

\section{PIDL-Based TSE on NGSIM Data}
\label{sec-VI}

This section evaluates the PIDL-based TSE method using real traffic data, the  Next  Generation  SIMulation (NGSIM) dataset\footnote{www.fhwa.dot.gov/publications/research/operations/07030/index.cfm}, and compares its performance to baselines.

\subsection{NGSIM Dataset}

NGSIM dataset provides detailed information about vehicle trajectories on several road scenarios. We focus on a segment of the US Highway 101 (US 101), monitored by a camera mounted on top of a high building on June 15, 2005. The locations and actions of each vehicle in the monitored region for a total of around 680 meters and  2,770 seconds were converted from camera videos. This dataset has gained a great attention in many traffic flow studies~\cite{Chiabaut-2010,Laval-2010, Montanino-2013, Fan-2013}.

\begin{figure}[h!]
\centering
  \includegraphics[scale=0.58]{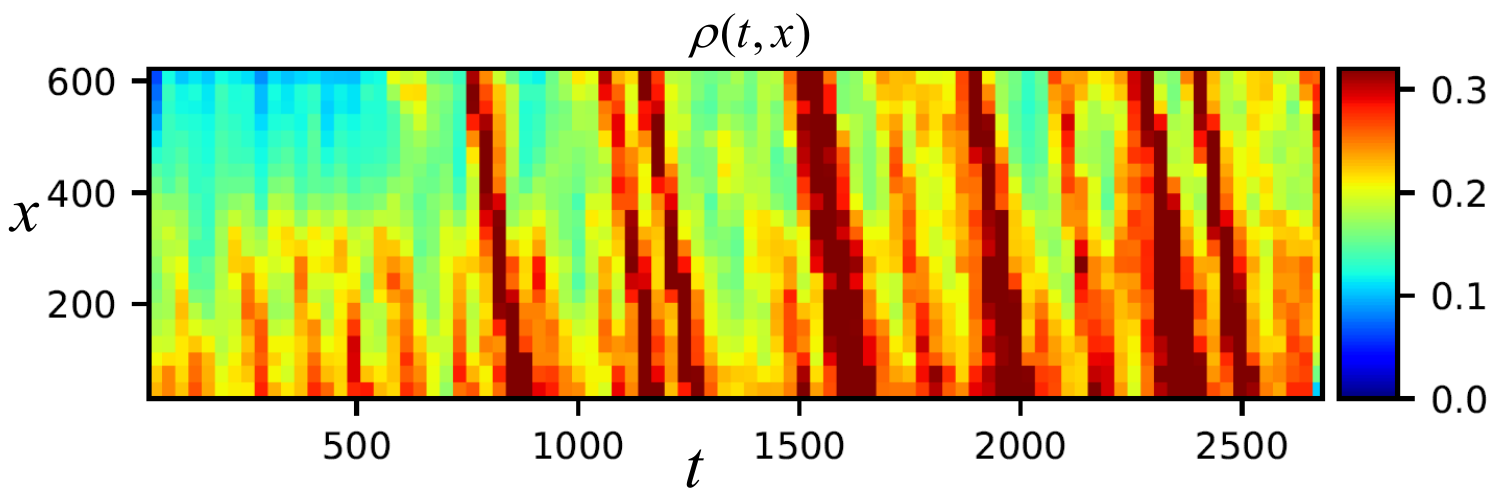}
  \caption{ Visualization of the average  traffic density on US 101 highway.}
  \label{fig:ch6-exact}
\end{figure}

\textcolor{black}{We select the data from all the mainline lanes of the US 101 highway segment to calculate the average traffic density for approximately every 30~meters over a 30~seconds period. After preprocessing to remove the time when there are non-monitored vehicles running on the road (at the beginning and end of the video), there are 21 and 89 valid cells on the spatial and temporal dimensions, respectively. We treat the center of each cell as a grid point. Fig.~\ref{fig:ch6-exact} shows the spatio-temporal field of traffic density in the dataset. From the figure, we can observe that shockwaves backpropagate along the highway.
}

For TSE experiments in this section, loop detectors are used to provide observed data with a recording frequency of 30 seconds. By default, they are evenly installed on the highway segment. Since no ground-truth model parameters are available for NGSIM, we skip the parameter discovery experiments.

\subsection{TSE Methods for Real Data}

We first introduce the PIDL-based method for the real world TSE problem, and then, describe the baseline TSE methods.

\subsubsection{PIDL-based Method}
This method is based on a three-parameter-based LWR model introduced by~\cite{Fan-2013}, which is shown to be more accurate in describing the NGSIM data than the Greenshields-based LWR. To be specific, their model sets the diffusion coefficient $\epsilon$ to $0$, and the traffic flow becomes $\rho_t+(Q(\rho))_x=0$, with the three-parameter flux $Q(\rho)$ defined in Eq.~(\ref{equ-13}). The PIDL architecture is Fig.~\ref{fig:ch5-PINN} is used. Because this is not a ring road, no boundary conditions are involved.

\begin{figure}[h!]
\centering
  \includegraphics[scale=0.52]{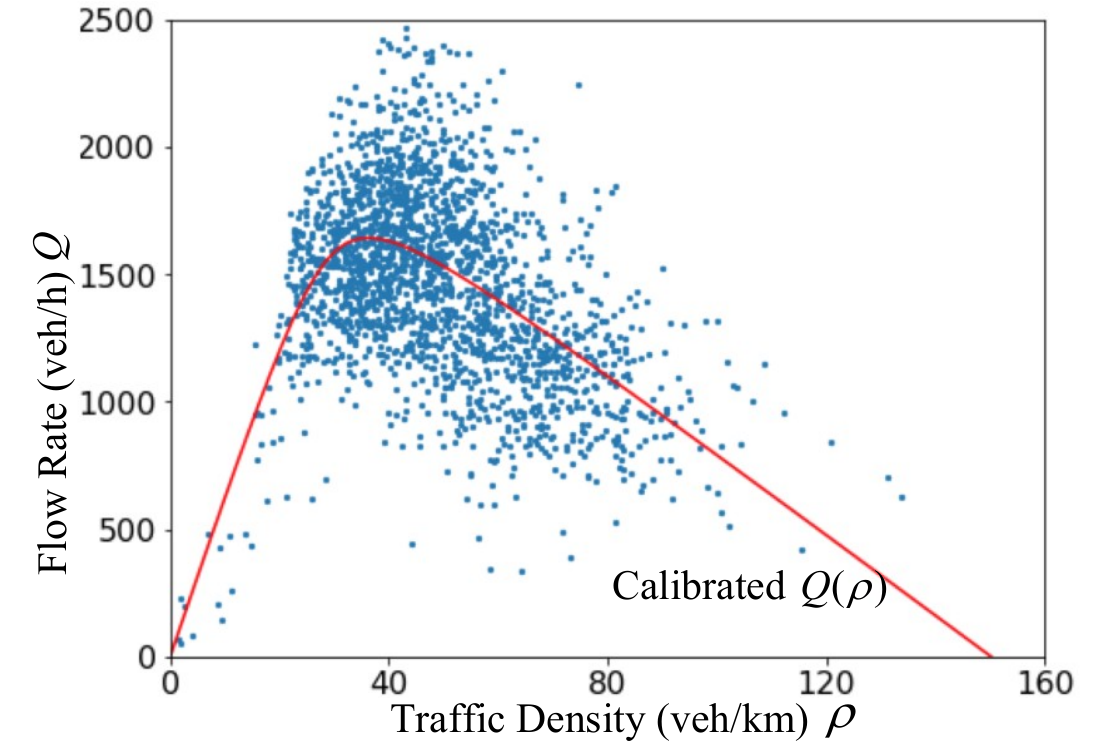}
  \caption{Calibration of the three-parameter flux $Q(\rho)$ using data (blue dots).}
  \label{fig:ch5-FD}
\end{figure}

The three-parameter flux $Q(\rho)$ is calibrated using the data from loop detectors, which measure the traffic density $\rho$ (veh/km) and flow rate $Q$~(veh/h) over time. Specifically, as shown in Fig.~\ref{fig:ch5-FD}, we use fundamental diagram (FD) data (i.e., flow-density data points) to fit the three-parameter flux function (an FD curve). Least-squares fitting~\cite{Fan-2013} is used to determine model parameters $\delta$, $p$, $\sigma$, and $\rho_{max}$. The calibrated model is then encoded into PIDL.

We select $N_a = 1,566$ auxiliary points $A$ from the grid~$G$ (80\%). The loss in Eq.~(\ref{equ-3-3}) (with boundary terms removed) is used for training the PUNN using the observed data from loop detectors (i.e., both observation points $O$ and corresponding target density values $P$). After hyperparamter tuning, we present the minimal-achievable estimation errors. The same for other baselines by default.

\subsubsection{Neural Netowrk (NN) Method}
This baseline method only contains the PUNN component in Fig.~\ref{fig:ch5-PINN}, and uses the first term in Eq.~(\ref{equ-3-3}) as the training loss. To be specific, we have $Loss_{\theta}=\alpha \cdot MSE_o$ for this baseline model and optimize the neural network using observed data for data consistency.

\subsubsection{Long Short Term Memory (LSTM) based Method}
This baseline method employs the LSTM architecture, which is customized from the LSTM-based TSE proposed by~\cite{LiWei-2018}. This model can be applied to our problem by leveraging the spatial dependency, i.e., to use the information of previous cells to estimate the traffic density of the next cell along the spatial dimension. We use the loss in Eq.~(\ref{equ-3-3}) to train the LSTM-based model for TSE.

\subsubsection{Extended Kalman Filter (EKF) Method}
This method applies the extended Kalman filter (EKF), which is a nonlinear version of the Kalman filter that linearizes the models about the current estimation~\cite{Kim-2018}. We use EKF to (1) estimate the whole spatio-temporal domain based on calibrated three-parameter-based LWR and (2) update its estimation based on the observed data from loop detectors.

\textcolor{black}{
To show the advantages of the proposed PIDL-based method, we limit our prior  knowledge about the traffic flow to LWR and investigate that on top of this physical model, how well PIDL can take advantage of it. Under this setting, the TSE baselines we designed are either model-free (e.g., NN, LSTM) or LWR-based (e.g., EKF). TSE methods related to other traffic flow models, such as PW, ARZ and METANET will be left for future research.}

\subsection{Results and Discussion}

\begin{figure}[h!]
\centering
  \includegraphics[scale=0.9]{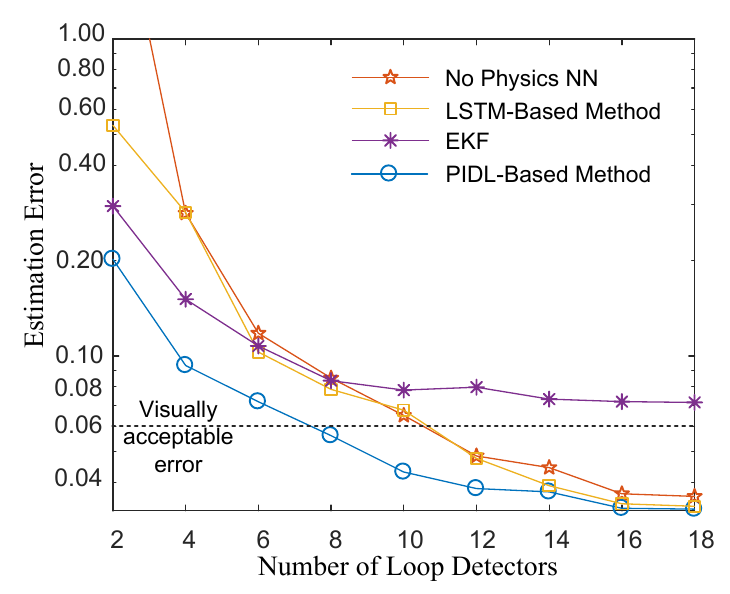}
  \caption{Comparison among PIDL-based methods and baseline methods.}
  \label{fig:ch6-resutls}
\end{figure}

We apply PIDL-based and baseline methods to TSE on the NGSIM dataset with different numbers of loop detectors. The results are presented in Fig.~\ref{fig:ch6-resutls}.

\textcolor{black}{From Fig.~\ref{fig:ch6-resutls}, we can observe that the PIDL-based method can achieve satisfactory estimation (with an error below $0.06$) using eight loop detectors. In contrast, other baselines need twelve or more to reach acceptable errors. The EKF method performs better than the NN/LSTM-based methods when the number of loop detectors is $<=6$, and NN/LSTM-based methods outperform EKF when sufficient data are available from more loop detectors. 
}

\textcolor{black}{The results are reasonable as EKF is a model-driven approach, making sufficient use of the traffic flow model to appropriately estimate unobserved values when limited data are available. However, the model cannot fully capture the complicated traffic dynamics in the real world, and as a result, the EKF's performance flattens out. NN/LSTM-based methods are data-driven approaches which can make ample use of data to capture the dynamics. However, their data efficiency is low and large amounts of data are needed for accurate TSE. The PIDL-based method's errors are generally below the baselines, because it can make efficient use of both the traffic flow model and observed data.
}

\begin{figure}[h!]
\centering
  \includegraphics[scale=0.75]{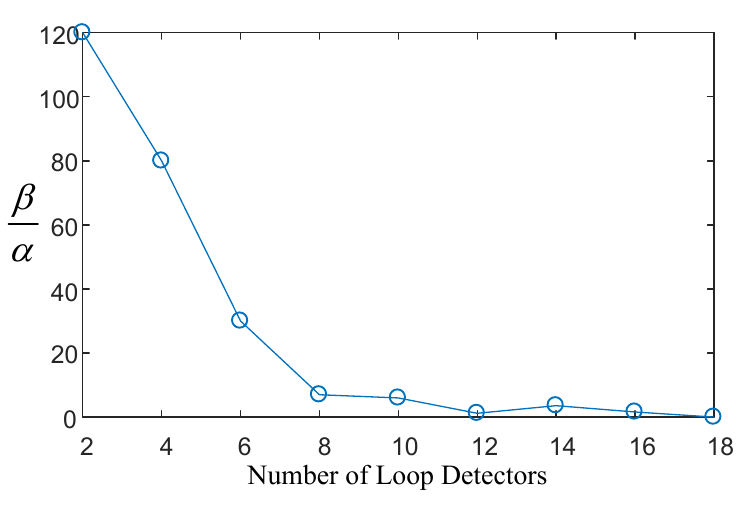}
  \vspace{-0.12in}
  \caption{Ratios of the contributions made by model-driven component and data-driven component to optimize the training process.}
  \label{fig:ch6-ratio}
\end{figure}

In addition to data efficiency, PIDL has the advantage to flexibly adjust the extent to which each of the data-driven and model-driven components will affect the training process. This flexibility is made possible by the hyperparamters $\alpha$ and $\beta$ in the loss Eq.~(\ref{equ-3-3}). Fig.~\ref{fig:ch6-ratio} shows the $\beta / \alpha$ ratios corresponding to the minimal-achievable estimation errors of the PIDL method presented in Fig.~\ref{fig:ch6-resutls}. When the number of loop detectors is small, more priorities should be given to the model-driven component (i.e., larger ratio)  as the model is the only knowledge for making the estimation generalizable to the unobserved domain. When sufficient data are made available by a large number of loop detectors, more priorities should be given to the data-driven component (i.e., smaller ratio) as the  traffic flow model model could be imperfect and can make counter effect to the estimation.

\begin{figure}[h!]
\centering
  \includegraphics[scale=0.90]{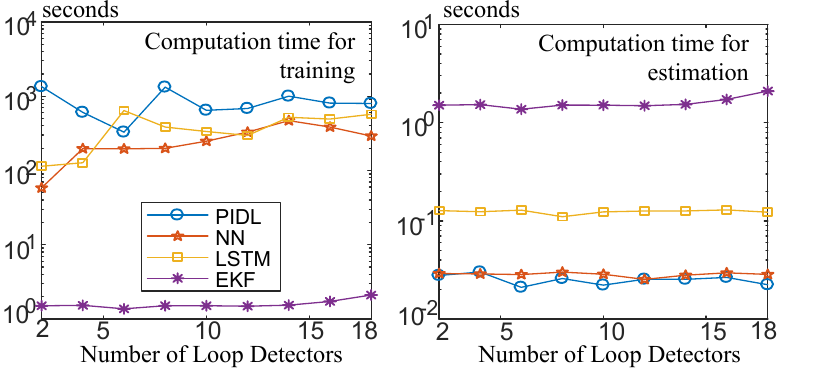}
  \vspace{-0.12in}
  \caption{Computation time for training and estimation of PIDL and baselines.}
  \label{fig:ch6-runtime}
\end{figure}

\textcolor{black}{Computation time for training and estimation is another topic that TSE practitioners may concern. The results of computation time corresponding to Fig.~\ref{fig:ch6-resutls} are presented in Fig.~\ref{fig:ch6-runtime}. Note that the training time and estimation time of EKF are the same because it conducts training and estimation simultaneously. For training, ML methods (NN and LSTM) consume more time than EKF because it takes thousands of iterations for ML methods to learn and converge, while EKF makes the estimation of the whole space per time step forward, which involves no iterative learning. PIDL takes the most time for training because it has to additionally backpropagete through the operations in PINN for computing the residual loss. For estimation, both PIDL and NN map the inputs directly to the traffic states and take the least computation time. LSTM operates in a recurrent way which takes more time to estimate. EKF needs to  update the state transition and observation matrices for estimation at each time step, and thus, consumes the most time to finish the estimation.
}

\section{Conclusion}

In this paper, we introduced the PIDL framework to the TSE problem on highways using loop detector data and demonstrate the significant benefit of using LWR-based  physics  to  inform  the  deep  learning. This framework can be used to handle both traffic state estimation and model parameter discovery. The experiments on real highway traffic data show that PIDL-based approaches can outperform   baseline methods in terms of estimation accuracy and data efficiency. Our work may inspire extended  research  on  using  sophisticated traffic  models for more complicated traffic scenarios in the future.

Future work is threefold: First, we will consider to integrate more sophisticated traffic models, such as the PW, METANET and ARZ models, to PIDL for further improving TSE and model parameter discovery. Second, we will explore another type of observed data collected by probe vehicles, and study whether PIDL can conduct a satisfactory TSE task using these mobile data. Third, in practice, it is important to determine the optimal hyperparameters of the loss given the observed data. We will explore hyperparameters finding strategies, such as cross validations, to mitigate this issue.

\section*{Acknowledgment}

The authors would like to thank Data Science Institute at Columbia University for providing a seed grant for this research.

\ifCLASSOPTIONcaptionsoff
  \newpage
\fi



%



\bibliographystyle{IEEEtran}

\bibliography{IEEEabrv, references.bib}

\vspace{5mm}

%







\end{document}